\documentclass[journal]{IEEEtran}

\ifCLASSINFOpdf
\else
\fi

\usepackage{times}
\usepackage{epsfig}
\usepackage{graphicx}
\usepackage{amsmath}
\usepackage{amssymb}
\usepackage{enumitem}
\usepackage{diagbox}
\usepackage{gensymb}
\usepackage{array}
\usepackage{color}
\usepackage{algorithm}
\usepackage{algpseudocode}
\makeatletter
\newcolumntype{L}[1]{>{\raggedright\arraybackslash}p{#1}}
\newcolumntype{C}[1]{>{\centering\arraybackslash}p{#1}}
\newcolumntype{R}[1]{>{\raggedleft\arraybackslash}p{#1}}

\begin{document}
%
\title{Multi-threshold Deep Metric Learning for Facial Expression Recognition}
%
%
%

\author{Wenwu~Yang,
        Jinyi~Yu,
        Tuo~Chen,
        Zhenguang~Liu,
        Xun~Wang,
        and~Jianbing~Shen,~\IEEEmembership{Senior Member,~IEEE}
\thanks{W. Yang, J. Yu, T. Chen, Z. Liu, and X. Wang are with the School of Computer and Information Engineering, Zhejiang Gongshang University, Hangzhou, 310018, China. (Email: wwyang@zjgsu.edu.cn).
}
\thanks{J. Shen is with the State Key Laboratory of Internet of Things for Smart City, Department of Computer and Information Science, University of Macau, Macau, China. (Email: shenjianbingcg@gmail.com).}
}

%
%

\markboth{}%
{Shell \MakeLowercase{\textit{et al.}}: Bare Demo of IEEEtran.cls for IEEE Journals}
%



\maketitle

\begin{abstract}
Effective expression feature representations generated by a triplet-based
deep metric learning are highly advantageous for facial expression recognition (FER).
The performance of deep metric learning algorithms
based on triplets is contingent upon identifying the best threshold for triplet loss. Threshold validation, however, is tough and challenging,
as the ideal threshold changes among datasets and even across classes within the same dataset.
In this paper, we present the multi-threshold deep metric learning technique, which not only avoids the difficult threshold validation but also vastly increases the capacity of triplet loss learning to construct expression feature representations.
We find that each threshold of the triplet loss intrinsically determines a distinctive distribution of inter-class variations and corresponds, thus, to a unique expression feature representation. Therefore,
rather than selecting a single optimal threshold from a valid threshold range, we thoroughly sample thresholds across the range, allowing the representation characteristics manifested by thresholds within the range to be fully extracted and leveraged for FER.
To realize this approach,
we partition the embedding layer of the deep metric learning network into a collection of slices and model training these embedding slices as an end-to-end multi-threshold deep metric learning problem.
Each embedding slice corresponds to a sample threshold and is learned by enforcing the corresponding triplet loss, yielding a set of distinct expression features, one for each embedding slice.
It makes the embedding layer, which is composed of a set of slices, a more informative and discriminative feature, hence enhancing the FER accuracy.
In addition, conventional triplet loss may fail to converge when using the popular \textit{Batch Hard} strategy to mine informative triplets.
We suggest that this issue is essentially a result of the so-called ``incomplete judgements'' inherent in the conventional triplet loss.
In order to address this issue, we propose a new loss known as dual triplet loss. The new loss is simple, yet effective, and converges rapidly. Extensive evaluations demonstrate the superior performance of the proposed approach on both posed and spontaneous facial expression datasets.
\end{abstract}


%
\IEEEpeerreviewmaketitle

\section{Introduction}
\begin{figure}[tb]
\centering
   \includegraphics[width=0.95\linewidth]{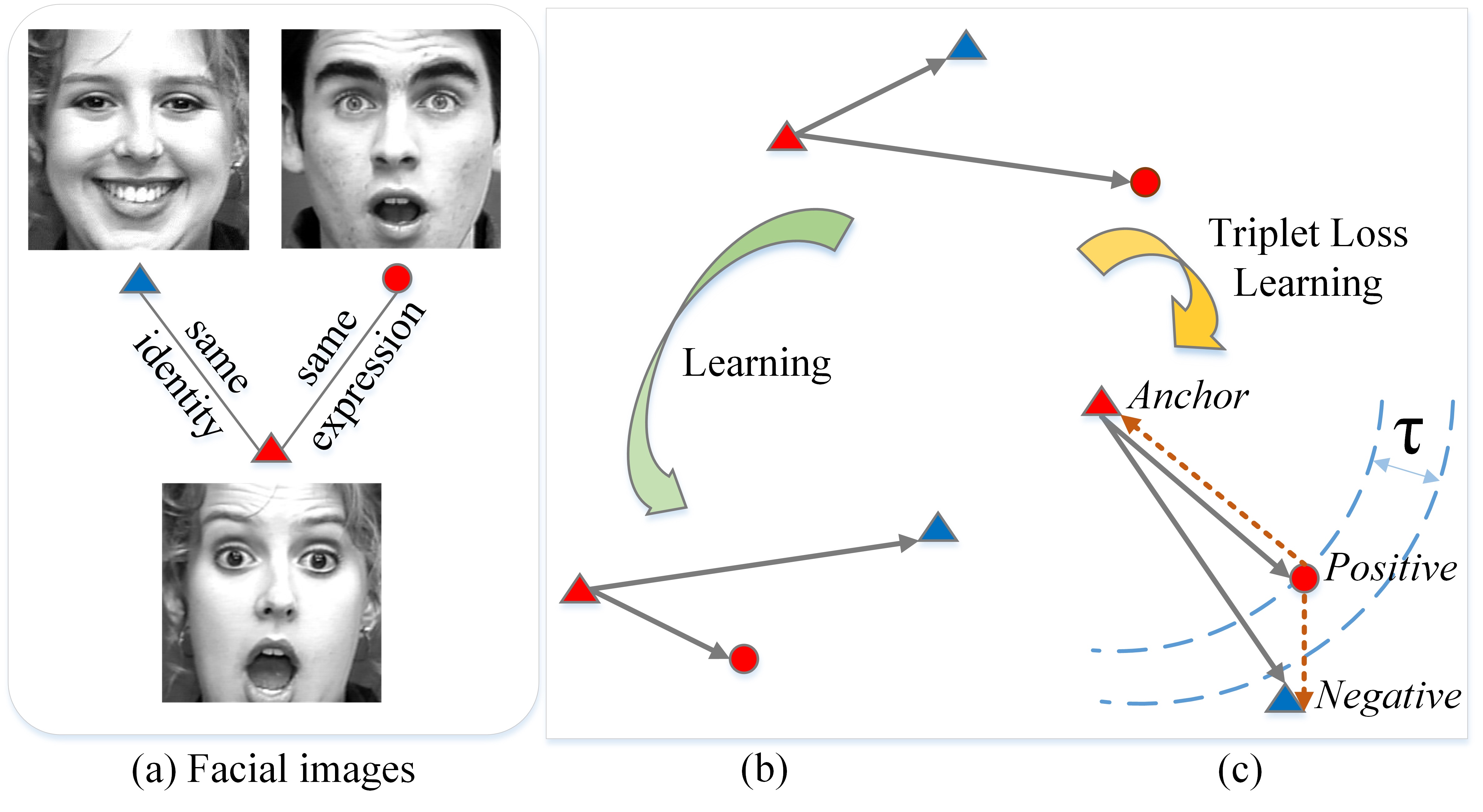}
   \caption{
   For face images (a), the feature distances may be dominated by the identity of the subjects rather than their expression (top of the right column), a problem that can be resolved with deep metric learning (b). In (c), the case of incomplete judgements is depicted, where the anchor is further from the positive than the negative (the dashed arrows), but the constraint imposed by the triplet loss has already been satisfied (the solid arrows), so incomplete judgements are not penalized. }
\label{fig:learing_demo}
\vspace{-4mm}
\end{figure}

%
%
%
%
\label{section:introduction}
\IEEEPARstart{F}{acial} expressions are one of the most expressive, natural, and universal ways for humans to communicate their emotional state.  Many studies on automatic facial expression recognition (FER) have been conducted due to its importance in a variety of applications such as human-machine interaction and health care~\cite{Li18}.
Despite significant efforts to improve recognition accuracy, FER remains a challenge due to variations in identity, pose, and illumination from face images~\cite{Liu17,BADDAR2020}.
For example, image features may be dominated more by identity variations than by facial expressions; as a result, as shown in Fig.~\ref{fig:learing_demo}, feature distances between different subjects with the \emph{same expression} may be larger than those between \emph{different expressions} of the same subject, implying that facial expression classification is not robust to image features.
Therefore, an effective expression feature representation is critical for increasing expression discrimination power. As shown in Fig.~\ref{fig:learing_demo}b, a deep metric learning scheme has the potential to learn from training data the latent features that effectively represent facial expression variations.

Deep metric learning for FER, in general, attempts to construct an expression feature representation defined by an embedding $f(x)$ , which maps an image $x$ into a feature space $\mathbb{R}^d$.
In this space, the feature distances between all face images with the same expression, regardless of imaging conditions or identities, are small, whereas the feature distances between face images with different expressions are large.
Previous deep metric learning methods based on image pairs \cite{Chopra05}, triplets \cite{Weinberger2009,Schroff15}, quadruplets \cite{Law2013,Li2017}, or $N$-tuplets \cite{Sohn16,Yu2019} can be used to learn such an embedding.
This work focuses on triplet-based deep metric learning because it is simple, easy to implement, and widely utilized in the community~\cite{Hermans17}.
A triplet is made up of a query image (\emph{anchor}), an image with the same expression as the query (\emph{positive}), and an image with a different expression than the query (\emph{negative}). In triplet-based deep metric learning, a triplet loss is imposed on the embedding $f(x)$ to ensure that the positive is closer to the anchor than the negative by a large margin $\tau$ \cite{Weinberger2009}, as illustrated in Fig.~\ref{fig:learing_demo}c.
In practice, a threshold validation process for identifying the optimal margin $\tau$, whether manual~\cite{Schroff15,Hermans17} or automatic~\cite{Liu17,Ge2018}, is required for triplet-based deep metric learning to achieve high performance.
However, threshold validation is extremely difficult because the optimal threshold $\tau$ changes between datasets and even classes within the same dataset~\cite{Ge2018,Wang2017}.

In this paper, we present an end-to-end learning approach, called the multi-threshold deep metric learning, which not only avoids the challenging threshold validation, but also greatly improves the ability of triplet loss learning to build expression feature representations.
Our theory is based on the observation that the threshold $\tau$ is usually valid within a range in which each threshold intrinsically specifies a different
distribution of inter-class variations and thus corresponds to a distinct expression feature representation.
As a result, rather than choosing only the optimal threshold from a valid range, we sample thresholds comprehensively over the range, allowing us to fully extract and leverage the representation characteristics manifested by thresholds within the range for FER.
In the proposed method, the embedding layer of a CNN-based deep neural network is divided into multiple non-overlapping slices, and each slice is formulated as a separate triplet loss learning network atop a shared CNN feature.
By assigning each sample threshold to one of the embedding slices, a collection of distinguishable expression features is learned, one for each embedding slice. As a result of composing its slices, a more diverse and discriminative feature is formed at the embedding layer, increasing the FER accuracy.

The proposed multi-threshold deep metric learning technique introduces a completely new viewpoint to triplet loss learning by assuming that each threshold within a valid range represents a unique representation characteristic that is distinct from those of the other thresholds; thus, it aims to fully extract and exploit the distinguishing characteristics manifested by thresholds within the range.
This theory differs substantially from those underlying previous approaches~\cite{Weinberger2009,Schroff15,Hermans17,Ge2018}, which seek to identify an optimal threshold within the valid range and assume that the feature learned by the optimal threshold can outperform and represent those learned by the others in the range.
Furthermore, despite the fact that conventional triplet loss may fail to converge
when the commonly used \textit{Batch Hard} strategy is employed to sample informative hard
triplets~\cite{Schroff15,Hermans17}, no reasonable explanation had been provided.
We suggest that this issue is essentially a result of the ``incomplete judgements'' inherent in the conventional triplet loss, as illustrated in Fig.~\ref{fig:learing_demo}c. In such cases, the distance between the anchor and the \textbf{positive} is greater than the distance between the negative and the \textbf{positive}, but the triplet loss constraints have already been satisfied, so the incomplete judgements are not penalized.
To address this problem, we propose the dual triplet loss, which effectively avoids the incomplete judgements and leads to rapid convergence.

Our main contributions are four-fold:
\begin{itemize}
  \item We propose the multi-threshold deep metric learning technique, which not
only avoids the challenging threshold validation, but also allows
the representation characteristics manifested by thresholds within a valid
range to be fully extracted and leveraged for FER, hence enhancing the performance of FER.
 \item
We build the proposed multi-threshold deep metric learning technique within the standard triplet loss learning framework, avoiding additional computational complexity and enabling the entire network to be trained end-to-end, such that such a technique can be employed in the existing triplet-based deep metric learning frameworks in a plug-and-play fashion.
\item We identify the issue of incomplete judgments and propose a simple but effective dual triplet loss as a solution, thereby achieving fast convergence in triplet loss learning.
\item Our proposed approach successfully improves performance on both posed and spontaneous facial expression databases by a considerable margin, compared to state-of-the-art methods.
\end{itemize}

\section{Related Work}
Facial expression recognition (FER) categorizes expressions into six or seven basic emotions that are considered independent of culture \cite{Matsumoto1992,Ramprakash2018}.
Automatic FER is a well-studied problem, and we refer interested readers to \cite{Kumari15,Li18} for a thorough examination.
According to the feature representations, FER methods can be classified into two categories: image-based \cite{Liu17,Meng17,Yang_2018_CVPR,Wang2020} and sequence-based \cite{Zhang-TIP2017,Kumawat2019,Perveen2020,Chen-TAC2020}. Our approach belongs to the image-based.

\begin{figure*}[tb]
\centering
   \includegraphics[width=1.0\linewidth]{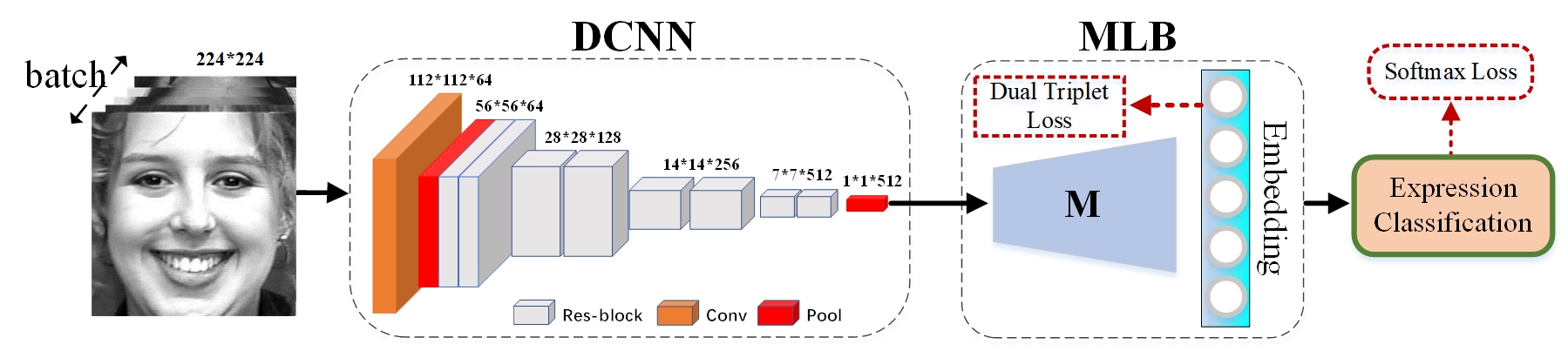}
   \caption{\label{fig:network}
The proposed deep metric learning network for FER. It consists of a batch input layer, a deep CNN (ResNet-18 in our implementation), an embedding layer, and a classification layer. Through a linear layer comprised of an embedding matrix $\mathbf{M}$, an embedding feature for the classification of expressions is learned in the embedding layer.}
 \vspace{-3mm}
\end{figure*}

\subsection{Expression feature representations}
An important issue on FER is how to extract expression features in facial images while ignoring expression-unrelated variations like illumination, head pose, and identity bias. Many handcrafted features have been proposed in the past, such as local binary patterns (LBP)~\cite{Shan2009}, LBP on three orthogonal planes (LBP-TOP)~\cite{zhao07}, or 3D-histogram of oriented gradients (3D-HOG)~\cite{Klaeser2008}.
However, discriminative feature representations learned by a CNN-based deep neural network have attained state-of-the-art performance and garnered an increasing amount of attention in recent years~\cite{Acharya18,Li18,Wang2020,She2021DiveIA}.
For example, a feature selection network is proposed in~\cite{Shuwen2018} to automatically extract and filter facial features for FER.  The de-expression learning procedure introduced in~\cite{Yang_2018_CVPR} uses a generative model of neutral face images to filter out expression information. Furthermore, \cite{Can2019} learns a robust expression feature representation by decoupling identity and pose information using adversarial learning. In~\cite{Bowen2019}, non-occluded images are used to guide the extraction of expression features from occluded cases.
In the meantime, several attempts have been made to address the ambiguity problem caused by subjective annotation and possible visually similarity between various facial expressions~\cite{WenXin_2020}. In~\cite{Zeng_2018_ECCV}, the end-to-end LTNet for discovering the latent truth from inconsistent pseudo labels and input face images is proposed. LDL-ALSG~\cite{Chen_CVPR2020} eliminates annotation
inconsistency through label distribution learning that leverages the topological information of the labels from additional tasks, including action unit recognition and facial landmark detection. In~\cite{Wang2020}, a self-cure network (SCN) is employed to relabel incorrectly labeled samples in order to reduce the negative impact of ambiguous data.
DMUE~\cite{She2021DiveIA} utilizes multiple auxiliary branches to discover the label distribution, which is then used to adjust the learning focus in conjunction with the original annotations. Recently, a data augmentation technique known as MixAugment has been proposed to enhance the performance of FER~\cite{Psaroudakis2022}.
In all of these CNN-based FER methods, softmax loss is the most frequently employed classification loss function~\cite{Barros2020}.
The feature embeddings learned by classification loss may be meaningful, but they have no direct relationship to the practical expressions. In terms of underlying emotions, for instance, the softmax loss explicitly promotes neither intra-class nor inter-class separation in feature embeddings.

\subsection{Expression embedding by deep metric learning}
Deep metric learning uses CNN-based deep neural networks to map an image into a feature vector in an embedding space.
In this embedding space, semantically similar images are close together, whereas semantically dissimilar images are far apart. Consequently, deep metric learning has the potential to learn feature embeddings that model expression variations from training data, thereby enhancing the FER accuracy.
To obtain effective deep embedding models, a plethora of approaches had been explored in the deep metric learning community~\cite{Mahmut2019}. Most of these approaches focus on new metric learning losses~\cite{Kihyuk2016,Sungyeon2020,Chopra2005,Schroff15,Sen2017,Yu2019-tuplet}, mining of informative samples~\cite{Bucher2016b,Harwood2017,Byungsoo2020,Xun2020,Milbich2022,ChaoYuan2017}, or ensemble models~\cite{Wonsik2018,Opitz2020,Yuhui2017}.
These deep metric learning techniques are frequently used to improve FER performance. In IACNN~\cite{Meng17}, an expression-sensitive contrastive loss and an identity-sensitive contrastive loss are employed to enhance the robustness of the FER network.
Similarly, a generalized (N+M)-tuplet clusters loss function adapted from (N+1)-tuplet loss~\cite{Kihyuk2016} is proposed along with identity-aware hard-negative mining to enhance the performance of FER~\cite{Liu17}. Inspired by center loss~\cite{Yandong2016}, DLP-CNN~\cite{Li19} improves the discriminative power of expression features by preserving the locality closeness and maximizing the inter-class scatter.
Similarly, IL-CNN~\cite{Cai18} enhances center loss by incorporating an additional island loss in order to acquire a more discriminative expression feature.
AMSCNN~\cite{Li18_CenterLoss} proposes a multi-scale CNN with an attention mechanism for robust FER, where a regularized version of the center loss is used to penalize the distance between expression features.
In \cite{Farzaneh2020}, a discriminant distribution-agnostic loss (DDA loss) is introduced to enforce inter-class separation of deep features for both majority and minority classes in order to learn better expression embedding in extreme class imbalance scenarios.
In this work, we vastly increase the capacity of triplet loss learning to construct expression feature
representations.
In contrast to previous methods, which either introduce more complex losses, additional multi-branch computations, or a complex mining strategy, our method strictly adheres to the standard triplet loss learning paradigm and is conceptually straightforward, simple to implement, and does not require additional computational complexity.

\subsection{Triplet loss-based embedding learning}
This work focuses on triplet-based deep metric learning to improve FER performance because it is simple, easy to implement, and widely used in the community~\cite{Hermans17}. The underlying principle of triplet loss is to ensure that, in the feature space, the distances between semantically dissimilar images are greater than the distances between semantically similar images by a large threshold $\tau$. In~\cite{Wang2017,Ge2018}, it has been established that a threshold validation procedure for determining the ideal threshold within a valid threshold range is necessary for triplet loss learning in order to achieve high performance. While human tweaking for the ideal threshold is laborious and time-consuming~\cite{Schroff15,Hermans17}, several approaches attempt to automatically discover the optimal threshold for each triplet or the triplet clusters~\cite{Liu17,Ge2018}. In this work, we introduce a completely new viewpoint to triplet loss
learning, namely that each threshold within a valid range
represents a unique representation characteristic that is distinct from
those of the other thresholds. Rather than
choosing only the optimal threshold from a valid range, we
sample thresholds comprehensively over the range, allowing
the representation characteristics
manifested by thresholds within the range to be fully extracted and leveraged for FER. In addition, the challenging threshold validation is not required by our method.

\begin{figure*}[tb]
\centering
   \includegraphics[width=0.95\linewidth]{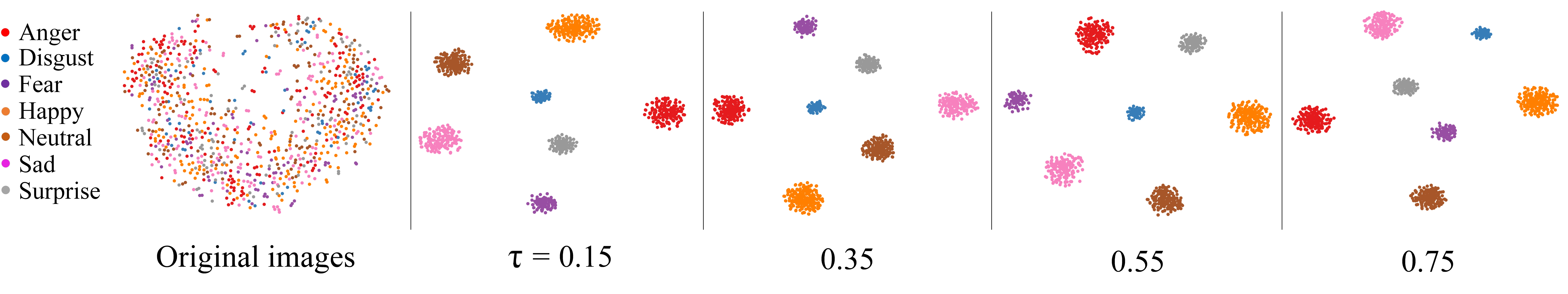}
   \caption{\label{fig:margin_demo}
  Distributions of inter-class variations with respect to the feature embeddings learned by triplet loss learning, where dual triplet loss is used as the loss function and the original images are samples from the facial expression database SFEW~\cite{Dhall15}. }
\end{figure*}

\section{Method}
Since our FER method is built on the triplet-based deep metric learning paradigm, we begin by revisiting triplet loss learning along with the motivation of our method. Then, in Section~\ref{sec:multilevel_deep_metric_fusion}, we describe the multi-threshold deep metric learning technique and its usage in FER in detail. Lastly, the dual triplet loss is introduced in Section~\ref{sec:symmetric_triplet_loss} to solve the issue of incomplete judgments.


\subsection{Triplet loss-based Embedding Learning for FER}
\label{sec:prelimiary}
An overview of the proposed triplet-based deep metric learning network for FER is illustrated in Fig.~\ref{fig:network}.
The network mainly consists of a deep CNN block (DCNN) and a metric learning block (MLB). The objective of the DCNN and MLB blocks is to learn a high-dimensional non-linear embedding $f(x)$ which maps an image $x$ into a feature space $\mathbb{R}^d$.
The DCNN block consisting of convolutional layers is trained to extract CNN features that are common to human face images, while a linear layer made up of an embedding matrix $\mathbf{M} \in \mathbb{R}^{h \times d}$ is learned in the MLB block to translate samples from the last hidden layer in the DCNN block of size $h$ into the feature space $\mathbb{R}^d$.
It is hoped that in this feature space, samples with the same expression will be drawn together, while samples with different expressions will be pushed apart.
To accomplish this, a triplet loss is used to constrain the embedding $f(x)$ when training the deep neural network.

Given a set of triplets, each consisting of an anchor $x_i^a$, a positive example $x_i^p$, and a negative example $x_i^n$, the triplet loss encourages the positive example to be closer to the anchor than
the negative example by a large margin $\tau$, i.e.,
\begin{equation}
\label{eqn:dis}
||f(x_i^n)-f(x_i^a)||_2 > ||f(x_i^p)-f(x_i^a)||_2 + \tau.
\end{equation}
Therefore, the standard triplet loss function is then $L_{triplet}=$
\begin{equation}
\label{eqn:loss_old}
\frac{1}{K}\sum_{i=1}^K\Big[||f(x_i^p)-f(x_i^a)||_2-||f(x_i^n) - f(x_i^a)||_2+\tau\Big]_+,
\end{equation}
where $[\cdot]_+ = max(0, \cdot)$ denotes the hinge function, and $K$ is the total number of triplets in the set.

A simple method for constructing triplets is to group triplets from training images; however, triplet samples will grow exponentially and become highly redundant, resulting in slow convergence and poor performance.
As a result, the \emph{Batch Hard} strategy is commonly used to mine informative triplets.
Specifically, for each sample $a$ in the batch, the hardest (i.e., farthest) positive and hardest (i.e., nearest) negative samples are chosen to form a triplet, with the sample $a$ serving as the \emph{anchor} and its hardest positive and negative samples serving as the \emph{positive} and \emph{negative} samples, respectively.

\subsection{Our Motivation}
\label{sec:threshold_analysis}
Let $g(\cdot)$ be the function specified by the DCNN block, which accepts an image $x$ as input and outputs the final CNN feature of size $h$, the embedding $f(x)$ can then be
\begin{equation}
\label{eqn:f}
f(x)= g(x)\mathbf{M},
\end{equation}
where $\mathbf{M} \in \mathbb{R}^{h \times d}$ is the embedding matrix corresponding to the linear layer in the MLB block. It is obvious that for various values of the margin $\tau$, the triplet loss constraint defined in Eqn.~(\ref{eqn:dis}) can be satisfied by the same embedding matrix (i.e., the same embedding) up to a different overall scale factor. It indicates that the resulting embedding $f(x)$ retains its intrinsic property and corresponds to an identical distribution of inter-class
variations when the margin $\tau$ alters.

However, similar to previous methods~\cite{Schroff15}, the linear layer in the MLB block is followed by $L_2$ normalization in our implementation in order to achieve training stability.
Thus, the embedding $f(x)$ is normalized into unit length such that the embedding of a margin cannot be simply scaled to satisfy the triplet loss constraints established using the other margins. Consequently, each threshold results in a feature embedding that has a separate distribution of inter-class variations and so corresponds to a distinct expression feature
representation. A demonstration is shown in Fig.~\ref{fig:margin_demo}, which motivates us to fully extract and leverage the representation characteristics manifested by margins within a valid threshold range, hence boosting the FER performance.

\subsection{Multi-threshold Deep Metric Learning for FER}
\label{sec:multilevel_deep_metric_fusion}

Given that the distribution of inter-class variations is intrinsically distinct for each threshold, triplet-based deep metric learning needs to identify the optimal threshold within a valid threshold range in order to achieve superior performance to the other thresholds, as done in the previous methods~\cite{Schroff15,Hermans17,Liu17,Ge2018}.
However, due to the fact that the
ideal threshold changes among datasets and even across classes within the same dataset, threshold validation is tough and challenging. Furthermore, such an optimal threshold strategy cannot fully extract and exploit the representation characteristics
manifested by thresholds within a valid threshold range.

To address the above issues, we propose the multi-threshold deep metric learning technique. In the multi-threshold deep metric learning, we execute a comprehensive threshold sampling over a specified valid threshold range and then apply triplet loss learning for each sample threshold to build a collection of distinct expression features. Thus, when a sufficiently dense threshold sampling is used, the representation characteristics manifested by thresholds within the range can be fully extracted and leveraged for FER. On the other hand, when nearby thresholds are close together, the distributions of inter-class variations of the adjacent thresholds tend to be similar, so that features learned by the neighboring thresholds are redundant. Therefore, given a valid threshold range of $[\tau_{min}, \tau_{max}]$,
we utilize an appropriate sampling interval $\delta\tau$ to sample this range evenly, resulting in a set of distinct sample thresholds, denoted as $\{\tau_i\}_{i=1}^N$:
\begin{equation}
\label{eqn:sampling}
\left\{
\begin{array}{cl}
   N=  &  (\tau_{max}-\tau_{min})/{\delta\tau}+1\\
   \tau_i=  & \tau_{min}+\delta\tau*(i-1)
\end{array}. \right.
\end{equation}

\begin{figure}[tb]
\centering
   \includegraphics[width=0.95\linewidth]{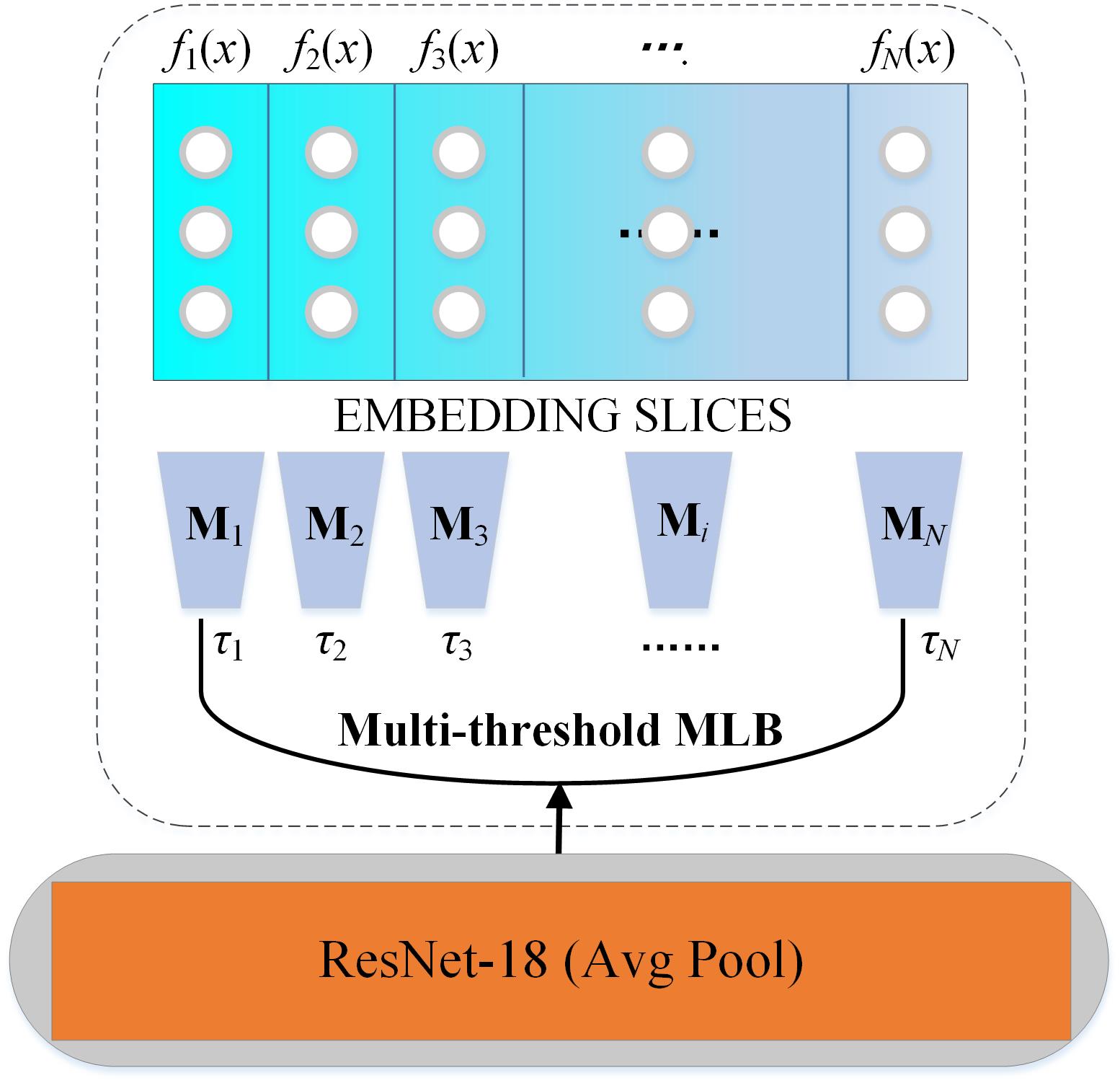}
   \caption{\label{fig:multi_network}
The module of the multi-threshold deep metric learning in which the embedding layer is divided into multiple non-overlapping slices and each slice is formulated as a separate triplet loss learning to produce a unique feature embedding $f_i(x)$ with a specific sample threshold $\tau_i$.}
\end{figure}

To fully exploit the representation characteristics manifested by each sample threshold for FER, a straightforward approach is to build an ensemble of $N$ deep CNN models, each of which is trained by triplet loss learning with a particular sample threshold. However, such a crude solution is too computationally intensive.
To circumvent this problem, we integrate the feature embeddings learned by each sample threshold into a single triplet loss learning network. In the meanwhile, we hope that the proposed approach is simple to implement and adheres to the triplet loss learning paradigm.
Keeping this in mind, our plan is to apply numerous triplet loss constraints on the embedding layer, each established with a specific sample threshold, as opposed to enforcing a single triplet loss constraint of the optimal threshold. To do so, we divide the embedding layer of the metric learning block (MLB) into $N$ fix-sized non-overlapping groups, as depicted in Fig.~\ref{fig:multi_network}, where each group is referred to as an embedding slice, and each sample threshold is assigned to one of these slices. Then, each embedding slice is formulated as a separate triplet loss learning network atop a shared CNN feature, which is extracted by a ResNet-18 CNN backbone in all our experiments. During training, each embedding slice is subject to a triplet loss defined by a specific sample threshold.

Though each embedding slice is formulated as a separate triplet loss learning network, denoted as $\{f_i(x)\}_{i=1}^N$, the key idea of the proposed multi-threshold deep metric learning technique is to learn these embedding slices in a unified and end-to-end manner. The network can be roughly divided into two parts, which correspond to the deep CNN block (DCNN) and the metric
learning block (MLB), respectively. In the DCNN block, the CNN features common to human facial images are extracted using a deep CNN network, such as a conventional ResNet-18 backbone in our experiment, where the final pooling layer of ResNet-18 results in a 512-dimensional CNN feature.
In the MLB block, for each embedding slice, a fully-connected layer made up of an embedding matrix $\mathbf{M}_i \in \mathbb{R}^{h \times d_i}$ is utilized to transform the $h$-dimensional CNN feature ($h=512$ in our experiment) into a $d_i$-dimensional embedding vector, i.e.,
\begin{equation}
\label{eqn:f_i}
f_i(x)= g(x)\mathbf{M}_i,
\end{equation}
where $g(\cdot)$ corresponds to the deep CNN network in the DCNN block. Hence, combining all the slices, i.e., concatenating all the $d_i$-dimensional embedding vectors at each slice, yields a $d$-dimensional embedding vector at the embedding layer, where $d = \sum_{i=1}^N d_i$, i.e., $f(x)=[f_1(x), f_2(x), \ldots, f_N(x)]$. Following the embedding layer, the expression classification result is generated via a  fully-connected layer.

In the training phase, each embedding slice is followed by a triplet loss $loss_i$, $i\in\{1, 2, \ldots, N\}$, defined by the corresponding sample threshold $\tau_i$, in order to learn the desired embeddings of $\{f_i(x)\}_{i=1}^N$. Consequently,
a set of unique expression features is generated, one for each embedding slice.
It makes the embedding layer, i.e., $f(x)$, which consists of a collection of slices, i.e., $f_i(x)$, $i\in\{1,2,\ldots,N\}$, a more informative and  discriminative feature, hence enhancing the FER performance.
All components of the proposed multi-threshold deep metric learning network are end-to-end jointly optimized, and the total loss function is defined as:
\begin{equation}
\label{eqn:total_loss}
{total\_loss} = \lambda\sum_{i=1}^N w_i*loss_i + loss_0,
\end{equation}
where $\lambda$ and $w_i$ are the weights and $loss_0$ is the softmax loss, which is defined as follows for expression classification:
\begin{equation}
\label{eqn:softmax_loss}
loss_0 = \frac{1}{M}\sum_{i=1}^M \bigg( -\sum_{c=1}^C { t_c^i \cdot \log\big(\frac{e^{y_c^i}}{\sum_{d=1}^C e^{y_d^i}} \big) } \bigg),
\end{equation}
where $M$ is the number of training images, $C$ is the number of expression classes, $[t_1^i, t_2^i, \ldots, t_C^i]$ is the one-hot encoding of the ground-truth expression class of a training image $x_i$, and $[y_1^i, y_2^i, \ldots, y_C^i]$ is the expression classification result of the image $x_i$, i.e., the final output of the network. The whole training procedure is illustrated in Algorithm~\ref{alg:myalgo}.
\begin{algorithm}
\caption{Multi-threshold deep metric learning algorithm.}\label{alg:myalgo}
\begin{algorithmic}
\State $M$=number of iterations, $N$=number of embedding slices
\For{$m=1$ \textbf{to} \texttt{$M$}}
    \State \texttt{/* \hspace{10pt} Forward Pass \hspace{10pt}   */}
    \State Let $\{x_k\}$ be the training samples in this batch
    \State $total\_{loss}=0$

    \For{$i=1$ \textbf{to} \texttt{$N$}}
        \State  Mine hard triplets from $\{x_k\}$ for the slice $f_i(x)$
        \State  Compute $loss_i$ by the dual triplet loss of Eq. (\ref{eqn:dual})
        \State  $total\_{loss} = total\_{loss} + w_i*loss_i$
    \EndFor
    \State Compute the softmax-loss $loss_0$ from $\{x_k\}$
    \State $total\_{loss} = {\lambda}\times{total\_{loss}} + loss_0$

    \State \texttt{/* \hspace{10pt} Backward Pass \hspace{10pt}   */}
    \State Update the network through back-propagation, where

    \State \hspace{28pt} $\frac{\partial loss_i}{\partial f_j} = 0$, when $i\neq j$ and $i\in\{1, 2, \ldots, N\}$
\EndFor
\end{algorithmic}
\end{algorithm}

\subsubsection{Weight tuning} Note that each embedding slice can be regarded as a separate triplet loss learning network, i.e., $f_j(x)$, $j\in\{1,2,\ldots,N\}$, where a specific triplet loss, i.e., $loss_i$, $i\in\{1, 2, \ldots, N\}$, is enforced on the slice during training. Since these embedding slices are non-overlapping, we have
\begin{equation}
    \frac{\partial loss_i}{\partial f_j} = 0, \hspace{8pt}\text{when}\hspace{4pt} i\neq j, i\in\{1, 2, \ldots, N\}.
\end{equation}
Therefore, the weights $\{w_i\}_{i=1}^N$ in Eqn.~(\ref{eqn:total_loss}), which are used to adjust the importance of each sample threshold on the embedding layer, are independent of one another. Consequently, the weight setting of $\{w_i\}_{i=1}^N$ is pointless, and we have set all of them to 1.
Thus, Eqn. (\ref{eqn:total_loss}) can be reduced as follows:
\begin{equation}
\label{eqn:total_loss2}
{total\_loss} = \lambda\sum_{i=1}^N loss_i + loss_0.
\end{equation}

\subsubsection{Analysis of computational complexity} In the multi-threshold deep metric learning network, the embedding layer is divided into a set of slices and each slice is formulated as a separate triplet loss learning network, i.e., $\{f_i(x)\}_{i=1}^N$. However, the overall computational complexity is the same as the original single triplet loss learning network, i.e., $f(x)$, which can be proved as follows. From Eqn.~(\ref{eqn:f_i}), we have $[f_1(x), f_2(x), \ldots, f_N(x)]=g(x)[\mathbf{M}_1, \mathbf{M}_2, \ldots, \mathbf{M}_N] $, where $\mathbf{M}_i \in \mathbb{R}^{h \times d_i}$. Since $f(x)= g(x)\mathbf{M}$ where $\mathbf{M} \in \mathbb{R}^{h \times d}$, and $d = \sum_{i=1}^N d_i$, it's easy to know that the total computational cost of $\{f_i(x)\}_{i=1}^N$ equals that of $f(x)$.

In fact, we can combine the embedding matrices of each embedding slices into a single one by setting $\mathbf{M}=[\mathbf{M}_1, \mathbf{M}_2, \ldots, \mathbf{M}_N]$. Therefore, in the inference time, the divided embedding slices can be ignored and the embedding layer can be viewed as a simple fully-connected layer made up of the embedding matrix of $\mathbf{M}$.

\begin{figure}[tb]
\centering
   \includegraphics[width=0.95\linewidth]{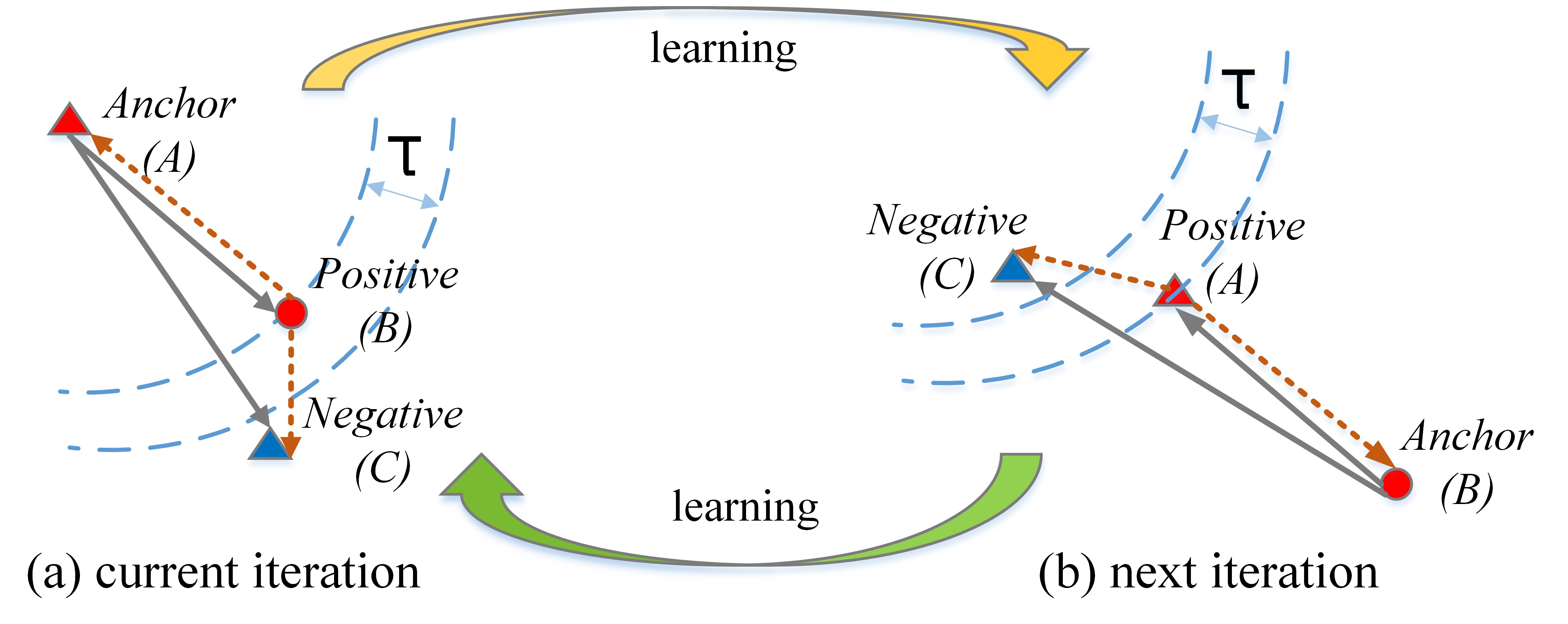}
   \caption{\label{fig:incomplete-cycle}
 Cycle of incomplete judgements during training time. In the cycle, the incomplete judgement of the hard triplet $(A, B, C)$ at the current iteration leads to the incomplete judgement of the hard triplet $(B, A, C)$ at the following iteration, and vice versa. }
\end{figure}

\subsection{Dual Triplet Loss}
\label{sec:symmetric_triplet_loss}
Similar to \cite{Schroff15,Hermans17}, we employ the \emph{Batch Hard} strategy to mine informative triplets for each embedding slice during the training phase, in order to speed up triplet loss learning and avoid inferior performance. However, conventional triplet loss may not converge. We
demonstrate that this issue is essentially a result of the so-called incomplete judgements inherent in the conventional triplet loss. Consider a hard triplet $(A, B, C)$, where $A$ is the anchor and $B$ and $C$ are its hardest positive and negative examples, respectively. A case of the incomplete judgement is illustrated in Fig.~\ref{fig:incomplete-cycle}a. In such a case, even though the triplet loss has pushed the positive example closer to the anchor than the negative example by a margin $\tau$ (note the solid arrows), the negative example unexpectedly becomes closer to the positive example than the anchor (see the dashed arrows). If $(B, A, C)$ is not a hard triplet in the current iteration, the case will not be penalized by the triplet loss. Even worse, $(B, A, C)$ may be selected as a hard triplet in the following iteration, and it is likewise handled by the incomplete judgement, as shown in Fig.~\ref{fig:incomplete-cycle}b. Consequently, the two triplets of $(A, B, C)$ and $(B,A,C)$ may be trapped in a cycle of incomplete judgements, causing triplet loss learning to fail to converge.

\begin{figure}[tb]
\centering
   \includegraphics[width=0.95\linewidth]{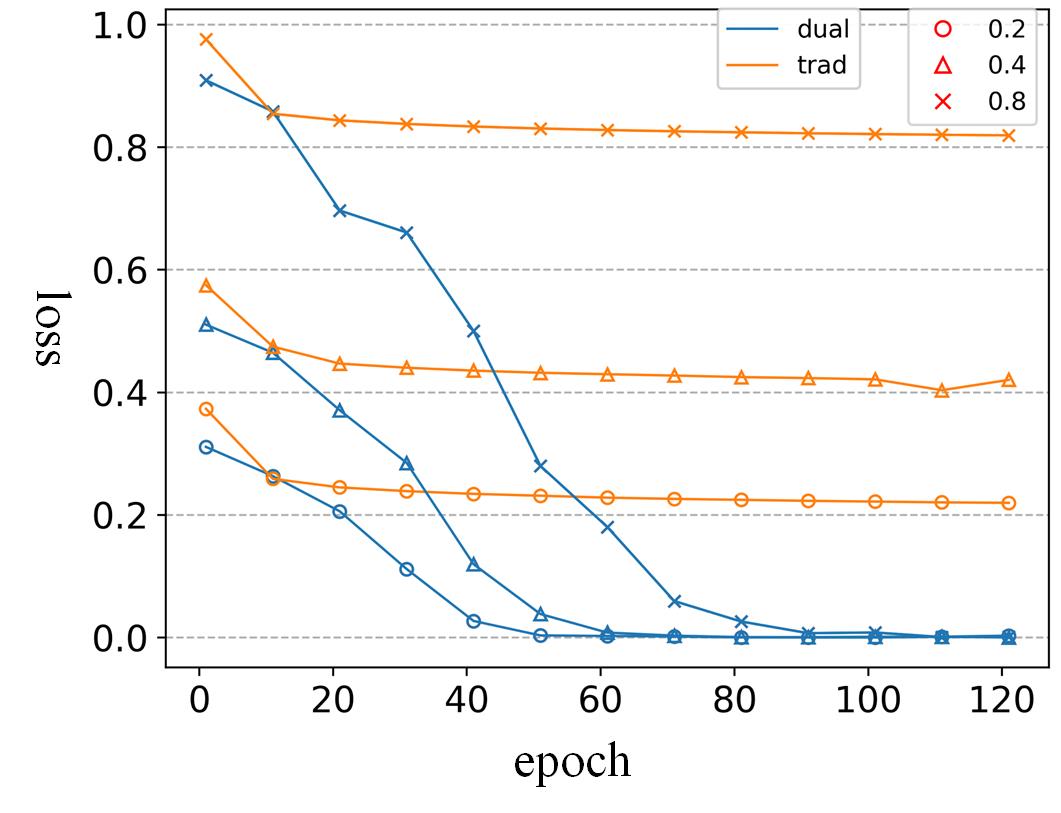}
   \caption{\label{fig:loss}
  The training loss from the SFEW database \cite{Dhall15}, utilizing three distinct thresholds ($0.2, 0.4, 0.8$) for both the traditional triplet loss (trad) and the dual triplet loss (dual). Note that validation loss is also shown in Fig.~\ref{fig:val_loss} of the Appendix~\ref{sec:validate_loss}.}
\end{figure}

To address the aforementioned problem, we introduce a new loss called dual triplet loss, i.e., $L_{dual}=$
\begin{equation}
\label{eqn:dual}
\begin{aligned}
&\frac{1}{2K}\sum_{i=1}^K\Big[||f(x_i^p)-f(x_i^a)||_2-||f(x_i^n) - f(x_i^a)||_2+\tau\Big]_+ \\
        &+\Big[ ||f(x_i^a)-f(x_i^p)||_2-||f(x_i^n)-f(x_i^p)||_2+\tau\Big]_+,
\end{aligned}
\end{equation}
where the second term is  essentially a repetition of the first and is used to penalize incomplete judgements by pushing the negative example further away from the positive example than the anchor, i.e.,
\begin{equation}
||f(x_i^n)-f(x_i^p)||_2 > ||f(x_i^a)-f(x_i^p)||_2 + \tau.
\end{equation}
The new loss can be Intuitively stated as follows: when a hard triplet of $(A, B, C)$ is constructed in the current iteration, the triplet of $(B, A, C)$ must also be defined as a hard triplet in the present iteration. This intuitive explanation also
also explains why the new loss is referred to as dual triplet loss. Clearly, the dual triplet loss prevents the possibility of incomplete judgments.
As shown in Fig.~\ref{fig:loss}, with the dual triplet loss, deep metric learning achieves rapid convergence, whereas the conventional triplet loss converges slowly.

\section{Implementation}
\subsection{Data Processing}
We use facial landmarks to crop the region of interest in the images and resize it to the size of $236\times236$. For databases that do not contain landmarks, the MTCNN~\cite{Zhang16_FaceDetect} is used for face detection and landmark localization. To prevent over-fitting, the data are augmented in two phases: offline and on-the-fly. In the offline phase, each image is rotated by $\{-10^{\degree}, -5^{\degree}, 0^{\degree}, 5^{\degree}, 10^{\degree} \}$, respectively.
During the training step, the input $224\times224$ patch samples are randomly cropped-out from five areas of the image (four corners and center, respectively) and then flipped horizontally to accomplish the on-the-fly augmentation.
The data augmentation is applied just to the training data, resulting in a $50$-times-larger dataset than the original. Only the central patch of the facial image is used for prediction during the inference phase.

\subsection{Optimization}
We employ the ResNet-18 in the DCNN block of the network and pre-train it using the facial emotion database of AffectNet~\cite{Mollahosseini2019}.  
We employ the Adam optimizer with a batch size of 96, a learning rate of 5e-6, and a dropout of 0.5 for the FC
layers during training. The network is trained end-to-end using $70$ epoches.
In Eqn.~(\ref{eqn:total_loss}) and~(\ref{eqn:total_loss2}) we set $\lambda=\frac{0.5}{N}$ for the triplet loss, where $N$ is the number of slices at the embedding layer. The widely used deep learning tool ``tensorflow \cite{Abadi16}'' is utilized to implement the network architecture.

\section{Experiments}
\label{sec:experiment}
We evaluate the proposed multi-threshold deep metric learning for FER (\mbox{Mul-DML}) on four well-known publicly available facial expression databases, including the laboratory-controlled databases of CK+ \cite{Lucey10} and MMI \cite{Pantic05} as well as the spontaneous expression databases of SFEW 2.0 \cite{Dhall15} and RAF-DB \cite{Li19}.
Comparatively, we employ two baseline methods: (1) CNN model, i.e., DCNN block + classification layer + softmax loss, and (2) \mbox{S-Dual-Triplet} model, i.e., the proposed triplet-based deep metric learning model for FER, as shown in Fig.~\ref{fig:network}, in which a dual triplet loss with the optimal threshold is enforced on the embedding layer and the embedding dimensionality is 256.

\subsection{Databases and Protocols}
\textbf{The Extended Cohn-Kanade database (CK+)} \cite{Lucey10} is a laboratory-controlled database widely used for evaluating FER. 327 video sequences from 118 subjects are labeled with one of seven expressions in CK+. As a standard protocol, the last three frames of each sequence are utilized, resulting in 981 images. We divide the images into 10 folds and conduct a 10-fold subject-independent cross-validation.

\textbf{The MMI database} \cite{Pantic05} contains 208 sequences captured from the frontal view of 31 subjects. They are labeled with six expressions (excluding "contempt"). 624 images are produced by utilizing three frames from the middle of each sequence where the expression reaches its peak. Similar to CK+, these images are divided into 10 folds for a subject-independent 10-fold cross validation.

\textbf{The SFEW database} \cite{Dhall15} is constructed by selecting frames from the AFEW database. SFEW 2.0, the most popular version, contains 1766 images, including Train (958), Val(436), and Test (372). Each image is assigned to one of the seven expressions, with the labels for the Train and Val sets being publicly accessible. Large variations in the SFEW reflect real-world conditions.

\textbf{The Real-world Affective Face Database
(RAF-DB)} \cite{Li19} is a relatively large ``in-the-wild'' database created by collecting 29672 extremely diverse facial images from the Internet. 15339 images are manually annotated with seven emotion categories, 12271 for training and 3068 for validation, using crowdsourcing.

\subsection{Ablation Studies}
We conduct ablation studies on the CK+, MMI, SFEW, and RAF-DB databases to validate the efficacy of individual components of our methodology.

\subsubsection{Impact of sampling interval and embedding dimensionality}
As demonstrated by Eqn.~(\ref{eqn:sampling}), in order to create sample thresholds $\{\tau_i\}_{i=1}^N$ from a given valid threshold range of $[\tau_{min}, \tau_{max}]$, an appropriate sampling interval $\delta\tau$ must be determined. Moreover, the dimensionality of embedding slices is required. Since both of them have an effect on the embedding layer and may have an effect on one another, we investigate both at the same time.
While a large margin is expected for triplet loss constraints, a margin that is too large will make it difficult for triplet loss learning to converge, as shown in Fig.~\ref{fig:loss}. In all experiments, we use the valid range of $[0.15, 0.75]$ so that examples from distinct classes can be effectively separated within $70$ epochs. Furthermore, since each embedding slice is formulated as a separate triplet loss learning network, we employ a fixed dimensionality for them.

\renewcommand{\arraystretch}{1.3}
\begin{table}
\begin{center}
\caption{Impacts of various sampling intervals and
embedding slice dimensions on the SFEW database~\cite{Dhall15}.}
\label{table:branch}
\begin{tabular}{| >{\centering}p{26pt} | >{\centering}p{50pt} | >{\centering}p{50pt} | >{\centering}p{50pt} |}
  \hline
   &  $\delta\tau=0.2$   &  $0.1$  & $0.05$ \tabularnewline
   \#dim & $(N=4)$ & $(N=7)$ & $(N=13)$ \tabularnewline
  \hline\hline
  512  &  $57.49\pm0.4$ &  $59.17\pm0.2$ &  $56.96\pm0.4$ \tabularnewline
  256  &  $57.65\pm0.8$ &  $60.09\pm0.3$ &  $57.65\pm0.7$ \tabularnewline
  128  &  $57.26\pm0.4$ &  $57.72\pm1.4$ &  $57.19\pm1.2$ \tabularnewline
  64  &  $55.81\pm1.0$ &  $55.96\pm0.6$ &  $58.10\pm0.9$ \tabularnewline
  \hline
\end{tabular}
\end{center}
\vspace{-5mm}
\end{table}

In Table~\ref{table:branch}, the effects of various sampling intervals and embedding slice dimensions are compared. One would expect that smaller sampling intervals and larger dimensions would result in improved performance. However, when small sampling intervals are utilized, e.g., $\delta\tau=0.05$, the distributions of inter-class variations of the adjacent thresholds tend to be
similar, so that features learned by the neighboring thresholds are redundant. Moreover, a small interval and a large dimension will result in more embedding slices and a larger network, which is computationally costly and more likely to over-fit. On the other hand, large intervals, such as $\delta\tau=0.2$, will result in insufficient sample thresholds, meaning that representation characteristics manifested by thresholds within the range cannot be fully extracted, resulting in degraded performance.  Overall, the sampling interval of $\delta\tau=0.1$, which results in seven sample thresholds, namely $\{0.15, 0.25, 0.35, 0.45, 0.55, 0.65, 0.75\}$, achieves the best performance. Accordingly, in all experiments we choose $\delta\tau=0.1$ and 256 for the dimension of embedding slices.

\subsubsection{Impact of different thresholds}
\label{sec:one-metric}
In Table~\ref{table:ablation_1Metric}, we compare triplet-loss learning models (\mbox{S-Dual-Triplet}) with varying thresholds. We can observe that the performance varies based on the threshold $\tau$, and the optimal threshold varies across databases. This difference in performance empirically verifies our analysis in section~\ref{sec:threshold_analysis} that distinct expression features could be learned for FER by varying threshold. In addition, it is shown in Table~\ref{table:ablation_Importance} that even with the optimal threshold, the performance of \mbox{S-Dual-Triplet} baseline is consistently inferior to that of the proposed approach (\mbox{Mul-DML}) by a large margin, i.e., $1.98\%$, $7.24\%$, $2.75\%$ and $1.95\%$ on CK+, MMI, SFEW and RAF-DB, respectively. It clearly demonstrates the superiority of our multi-threshold deep metric learning technique for FER, which fully extracts and exploits the distinctive features manifested by thresholds within a valid threshold range, as opposed to relying solely on the feature learned by the optimal threshold within the range.

\renewcommand{\arraystretch}{1.3}
\begin{table}
\begin{center}
\caption{Comparison of the performance of \mbox{S-Dual-Triplet} models with various thresholds. }
\label{table:ablation_1Metric}
\begin{tabular}{p{26pt}|p{17pt}p{17pt}p{17pt}p{17pt}p{17pt}p{17pt}p{20pt}}
  \hline
  $\tau=$ & 0.15  &  0.25  &  0.35  &  0.45  &  0.55  &  0.65  &  0.75 \\
  \hline
  CK+ & 94.91 &  94.42 &  93.84 &  \textbf{96.51} &  94.61 &  96.11 &  94.03 \\
  MMI & 73.36  & 71.00 &  73.97 &  \textbf{74.33} &  72.36 &  72.91  & 73.78 \\
  SFEW & 56.88 &  56.42 &  54.36 &  56.88  & 56.65 &  56.42 &  \textbf{57.34} \\
  RAF & 85.36 &  85.40 &  85.36 &  85.20  & 84.84 &  \textbf{85.63} &  85.16 \\
  \hline
\end{tabular}
\end{center}
\end{table}

\begin{figure*}[tb]
\centering
\vspace{-3mm}
   \includegraphics[width=0.95\linewidth]{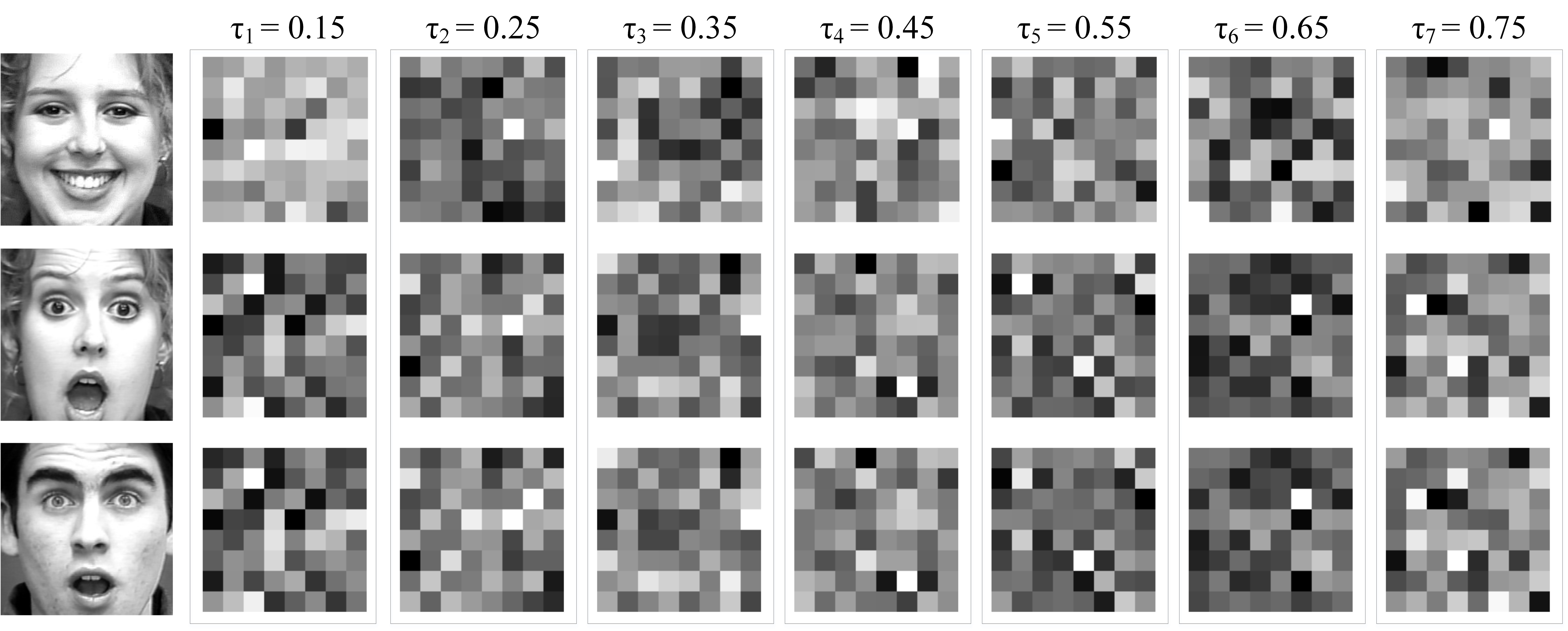}
   \caption{\label{fig:visualization}
  Examples of the acquired representation characteristics for each slice of the embedding. The first row is happiness, while the second and the third row are surprise. Each representation characteristic is visualized by the condition of neurons in the corresponding embedding slice, where
  each neuron corresponds to a dimension of the embedding slice. In each picture, each square represents a neuron and a brighter color indicates a greater value. For each embedding slice, 64 neurons are sampled equally from its 256 dimensions in order to display them clearly. Better viewed with zoom-in. }
\end{figure*}

\begin{figure*}
\centering
   \includegraphics[width=0.90\linewidth]{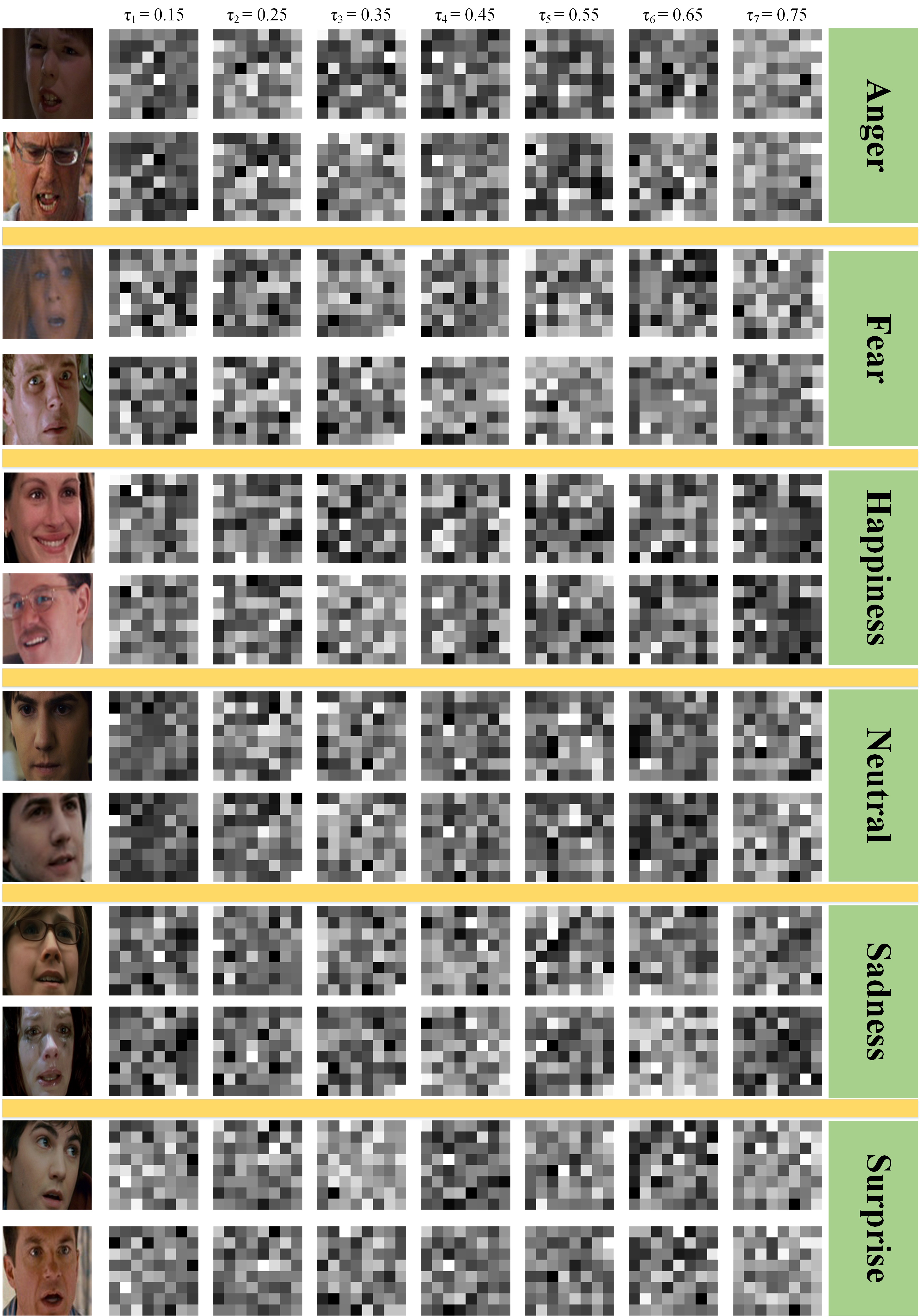}
   \caption{\label{fig:visualization2}
  More examples of the learned representation characteristics at each embedding slice. These examples include six expressions and are from the validation set of the SFEW database. Note that the features are rendered identically to those in Fig.~\ref{fig:visualization}. Better viewed with zoom-in.}
\end{figure*}

\subsubsection{Impact of the dual triplet loss in Mul-DML}
To quantify the contribution of dual triplet loss to the performance improvement of the proposed method, a new network, \mbox{S-Trad-Triplet}, was constructed. This network has the same structure as the \mbox{S-Dual-Triplet} baseline, but during training the traditional triplet loss rather than the dual triplet loss is enforced on its embedding layer. To ensure a fair comparison, more epochs were used to train this network until convergence. Table~\ref{table:ablation_Importance} presents the performance of the three networks: \mbox{S-Trad-Triplet}, \mbox{S-Dual-Triplet} and \mbox{Mul-DML}, where the optimal threshold from Table~\ref{table:ablation_1Metric} is applied to the first two networks. The dual triplet loss improves the accuracy of the CK+, MMI, SFEW and RAF-DB databases by $2.31\%$, $0.08\%$, $0.69\%$ and $0.14\%$, respectively, when compared to the traditional triplet loss. Moreover, compared to the dual triplet loss, the proposed multi-threshold deep metric learning technique can improve the performance of the four databases by a significant margin, namely $1.98\%$, $7.24\%$, $2.75\%$ and $1.95\%$, respectively.
Obviously, while the dual triplet loss has an undeniably positive effect, the multi-threshold deep metric learning improves the accuracy substantially. Consequently, we can conclude that the multi-threshold deep metric learning technique is primarily responsible for the enhanced FER performance of the proposed method.

\renewcommand{\arraystretch}{1.3}
\begin{table}
\vspace{-2mm}
\begin{center}
\caption{Performance improvements by dual triplet loss and multi-threshold deep metric learning technique.}
\label{table:ablation_Importance}
\begin{tabular}{| p{50pt} | >{\centering}p{30pt} | >{\centering}p{30pt} | >{\centering}p{30pt} | >{\centering}p{30pt} |}
  \hline
  Models & CK+  &  MMI  &  SFEW  & RAF-DB\tabularnewline
  \hline
  S-Trad-Triplet & 94.20 &  74.25 &  56.65 & 85.49 \tabularnewline
  S-Dual-Triplet & 96.51  & 74.33 &  57.34 & 85.63 \tabularnewline
  Mul-DML & 98.49 &  81.57 &  60.09 & 87.58 \tabularnewline
  \hline
\end{tabular}
\end{center}
\end{table}

\subsubsection{Impact of multi-thresholds and multi-slices in Mul-DML}
Note that the network structure of the proposed multi-threshold deep metric learning (\mbox{Mul-DML}) is identical to that of triplet-based deep metric learning, such as \mbox{S-Dual-Triplet}. The difference lies in two aspects: (1) the embedding dimensionality of \mbox{Mul-DML} is $256\times7$, whereas the embedding dimensionality of \mbox{S-Dual-Triplet} is $256$; (2) during the training phase, \mbox{S-Dual-Triplet} enforces a triplet loss on the embedding layer, whereas \mbox{Mul-DML} enforces a triplet loss on each slice in the embedding layer. To demonstrate the key factors in \mbox{Mul-DML}, we employ two additional training strategies on the \mbox{Mul-DML} network, resulting in two new models: \mbox{S-DML} and \mbox{Mul-DML-Same}. When training the \mbox{S-DML} model, only a triplet loss defined by the optimal threshold is enforced on the embedding layer of the \mbox{Mul-DML}'s network. Clearly, the \mbox{S-DML} model is comparable to the \mbox{S-Dual-Triplet} model, but with a larger embedding dimension, $256\times7$. In training \mbox{Mul-DML-Same}, a triplet loss is also enforced on each of the embedding slices, but each slice's triplet loss is defined by the same threshold, i.e., the optimal one in Table~\ref{table:ablation_1Metric}.

The upper portion of Table~\ref{table:ablation_multithreshold} depicts the different training strategies utilized by \mbox{S-DML}, \mbox{Mul-DML-Same}, and \mbox{Mul-DML}, whereas the lower portion provides the performance of the three models. Compared to \mbox{S-DML}, the \mbox{Mul-DML} model consistently improves the accuracy of the four databases by a substantial margin, i.e., (1.63\%, 6.91\%, 2.65\%, and 2.15\%).
In addition, Tables~\ref{table:ablation_Importance} and~\ref{table:ablation_multithreshold} reveal that the accuracy improvement of \mbox{S-DML}
over \mbox{S-Dual-Triplet} is very marginal. We can therefore conclude that the superiority of \mbox{Mul-DML} is primarily attributable to the multi-slice triplet loss learning strategy rather than the larger embedding dimensionality.
Moreover, we can see that \mbox{Mul-DML-Same} performs similarly to \mbox{S-Dual-Triplet} and \mbox{S-DML}, but consistently worse than \mbox{Mul-DML}, i.e., (-1.82\%, -5.94\%, -2.75\%, and -1.3\%) across the four databases. It demonstrates that a straightforward triplet loss training of multiple embedding slices, without the use of distinct thresholds, is insufficient for enhancing FER performance.

\begin{table}
\begin{center}
\caption{Performance comparison of various training strategies on the \mbox{Mul-DML} network }
 \label{table:ablation_multithreshold}
\begin{tabular}{| p{60pt} | >{\centering}p{55pt} | >{\centering}p{70pt} | }
  \hline
  Models & w/o multi-slices  &  w/o multi-thresholds \tabularnewline
  \hline
  S-DML & - &  - \tabularnewline
  Mul-DML-Same & \checkmark & - \tabularnewline
  Mul-DML & \checkmark & \checkmark \tabularnewline
  \hline
\end{tabular}
\vspace{2mm}
\end{center}
\begin{center}
\begin{tabular}{| p{60pt} | >{\centering}p{30pt} | >{\centering}p{30pt} | >{\centering}p{30pt} | >{\centering}p{30pt} |}
  \hline
  Models & CK+  &  MMI  &  SFEW  & RAF-DB \tabularnewline
  \hline
  S-DML & 96.86 &  74.66 &  57.44 & 85.43 \tabularnewline
  Mul-DML-Same & 96.67 & 75.63 & 57.34 & 86.28 \tabularnewline
  Mul-DML & \textbf{98.49} &  \textbf{81.57} &  \textbf{60.09} & \textbf{87.58} \tabularnewline
  \hline
\end{tabular}
\end{center}
\end{table}

\begin{figure*}[tb]
\vspace{-5mm}
\centering
   \includegraphics[width=1.0\linewidth]{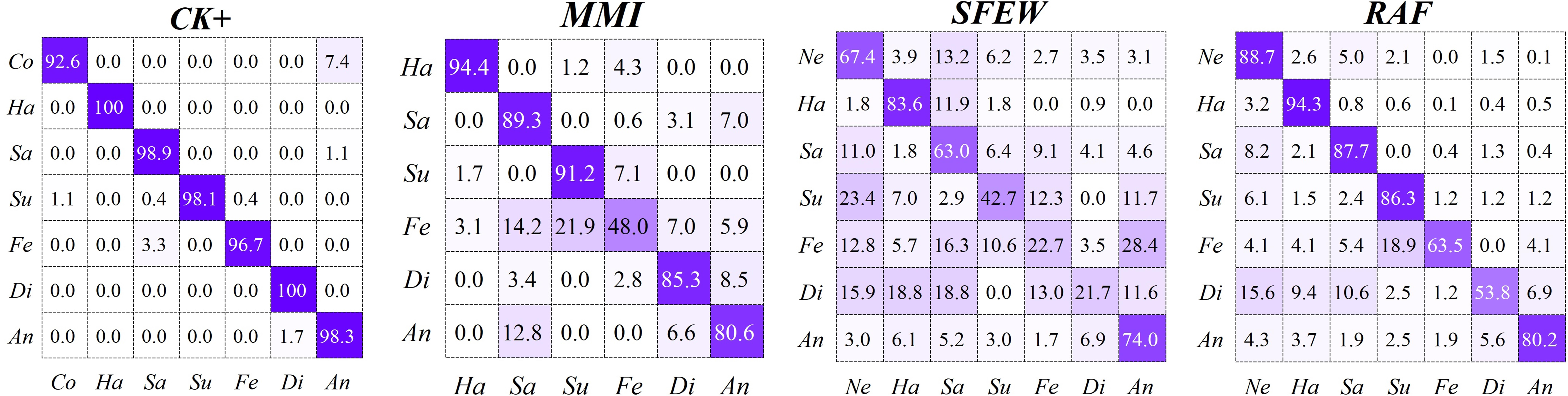}
   \caption{\label{fig:confusion_total}
   The confusion matrix on the laboratory-controlled databases of CK+ and MMI as well as the in-the-wild databases of SFEW 2.0 and RAF-DB.}
\vspace{3mm}
\end{figure*}

\subsection{Diversity of the Feature Representations}
Fig.~\ref{fig:visualization} depicts the expression features learned by each embedding slice during multi-threshold deep metric learning. More results are presented in Fig.~\ref{fig:visualization2}, wherein the examples are more diverse and exhibit substantial variations that reflect real-world conditions. The following intuitively consistent results can be observed in each of these cases. At each embedding slice, very similar features are constructed for different subjects with identical expressions, whereas features with large variations are generated for the same subject when he or she is experiencing different emotions. Consequently, we can conclude that every feature learned by an embedding slice is discriminative in terms of expression variations. Moreover, it is evident that, given a face image, the features generated by each embedding slice are notably distinct from one another. As a result, composing the features of all embedding slices, which are emotionally distinct and very dissimilar to one another, will result in a more diverse and discriminative expression feature at the embedding layer of the network, thereby enhancing the FER accuracy.

\subsection{Expression Recognition Evaluation}
In Tables~\ref{table:CK_Plus} and \ref{table:SFEW}, we compare our method to the baseline and state-of-the-art methods.
In consideration of their high relevance, we have highlighted the FER methods based on cutting-edge deep metric learning techniques by separating them and denoting them with an asterisk for clarity.
In addition, for comparative purposes, we have added two baseline models (marked with a plus sign) that have the same network structure as the \mbox{S-Dual-Triplet} model but are trained with the well-known Circle loss~\cite{Yifan2020} and the more recent HIST loss~\cite{Jongin2022}, respectively, instead of the dual triplet loss.

\subsubsection{The extended cohn-Kanade database (CK+)}
Table~\ref{table:CK_Plus} (third column) shows the average recognition accuracy of seven expressions across 10 runs on the CK+
database. Our method has a 98.49\% accuracy rate, which is the highest among the state-of-the-art approaches. Moreover, as a result of the proposed multi-threshold deep metric learning technique, our method consistently outperforms the baseline methods.
As depicted by the confusion matrix in Fig.~\ref{fig:confusion_total}, our method recognizes disgust and happiness extremely well, while contempt has the lowest recognition rate at 92.6\%. It may be the result of data acquisition in which the number of annotations indicating contempt is low.

\subsubsection{The MMI database}
The average recognition accuracy of six expressions over 10 runs on the MMI database is shown in Table~\ref{table:CK_Plus} (4th column). Our method achieves a recognition rate of over 81.5\%, outperforming the state-of-the-art and baseline methods by a considerable margin.
Our confusion matrix is depicted in Fig.~\ref{fig:confusion_total}, where fear has the lowest recognition rate and is most frequently confused with surprise. This is due to the fact that fear and surprise may appear visually similar.

\begin{table}
\begin{center}
\caption{Recognition accuracy on the laboratory-controlled databases of CK+ and MMI.}
 \label{table:CK_Plus}
\begin{tabular}{|R{85pt}|C{50pt}|C{30pt}|C{30pt}|}
  \hline
  Method & Setting & CK+ & MMI \\
  \hline\hline
  LBP-TOP \cite{zhao07} & sequence &  88.99 & 59.51\\
  HOG3D \cite{Klaeser2008} & sequence & 91.44 & 60.89\\
  STM-Explet \cite{Liu14}  & sequence & 94.19 & 75.12\\
  DTAGN \cite{Jung15}  & image &  97.25 & 70.24 \\
  DeRF \cite{Yang_2018_CVPR} & image & 97.30 & 73.23\\
  G2-VER \cite{Albrici2019} & sequence & 97.4 & N/A\\
  WS-LGRN \cite{Zhang-IJCAI2020} & image & \underline{98.37} & N/A\\
  LDL-ALSG \cite{Chen_CVPR2020} & image & 93.08 & 70.49 \\
  IF-GAN~\cite{Cai_Jie2021} & image & 97.52 & 75.48 \\
  \hline
  $\text{IACNN}^*$ \cite{Meng17} & image & 95.37 & 71.55 \\
  $\text{N+M}^*$ \cite{Liu17} & image & 97.1 & 78.53\\
  $\text{AMSCNN}^*$ \cite{Li18_CenterLoss} & image & 98.20 & N/A\\
  \hline
  CNN (baseline)  & image & 93.99 & 71.25  \\
  S-Dual-Triplet (baseline)  & image & 96.51  & 74.33 \\
  $\text{Circle Loss}^+$~\cite{Yifan2020}  & image & 96.54 & 76.42 \\
  $\text{HIST Loss}^+$~\cite{Jongin2022}  & image & 97.48 & \underline{79.13} \\
 \textbf{Mul-DML (Ours) } & image & \textbf{98.49}& \textbf{81.57}  \\
  \hline
\end{tabular}
\end{center}
 \vspace{-2mm}
\end{table}

\subsubsection{The SFEW database}
The SFEW database is diverse and has a wide range of pose and illumination, as demonstrated in the supplementary material, which also contains the recognition results for each expression class.

\begin{table}
\begin{center}
\caption{Recognition accuracy on the in-the-wild databases of SFEW and RAF-DB.}
 \label{table:SFEW}
\begin{tabular}{|R{85pt}|C{50pt}|C{30pt}|C{30pt}|}
  \hline
  Method & Setting & SFEW & RAF-DB \\
  \hline\hline
  LBP-TOP \cite{zhao07} & sequence & 25.13 & N/A \\
  HOG3D \cite{Klaeser2008} & sequence & 26.90 & N/A \\
  LTNet \cite{Zeng_2018_ECCV} & image &  \underline{58.29} & 86.77\\
  Cov. Pooling  \cite{Acharya18} & image & 58.14 & 87.0  \\
  SCN \cite{Wang2020} & image & N/A & 87.03 \\
  WS-LGRN \cite{Zhang-IJCAI2020} & image & N/A & 85.79\\
  LDL-ALSG \cite{Chen_CVPR2020} & image & 56.50 & 85.53 \\
  DMUE \cite{She2021DiveIA} & image & 57.12 & \textbf{88.76} \\
  MixAugment \cite{Psaroudakis2022} & image & N/A & \underline{87.54} \\
  \hline
  $\text{IACNN}^*$ \cite{Meng17} & image & 50.98 & N/A \\
  $\text{N+M}^*$ \cite{Liu17} & image & 54.19 & N/A\\
  $\text{IL-CNN}^*$ \cite{Cai18} & image & 52.52 & N/A\\
  $\text{DLP-CNN}^*$ \cite{Li19} & image & N/A & 84.13\\
  $\text{DDA}^*$ \cite{Farzaneh2020} & image & N/A & 86.90 \\
  \hline
  CNN (baseline) &  image & 54.58 & 84.94 \\
  S-Dual-Triplet (baseline)  & image  &  57.34 & 85.63  \\
  $\text{Circle Loss}^+$~\cite{Yifan2020}  & image & 57.11 & 86.02 \\
  $\text{HIST Loss}^+$~\cite{Jongin2022}  & image & \underline{58.30} & 87.22 \\
  \textbf{Mul-DML (Ours)} & image  & \textbf{60.09} & \underline{87.58} \\
  \hline
\end{tabular}
\end{center}
 \vspace{-3mm}
\end{table}

As shown in Table VI, column 3, our proposed method works well in real-world environments and achieves a 60.09\% recognition rate, which significantly outperforms the state-of-the-art methods. Note that the sequence-based traditional methods LBP-TOP~\cite{zhao07} and HOG3D~\cite{Klaeser2008} are tested on the AFEW database. Similar to the laboratory-controlled CK+ and MMI databases, our method consistently outperforms the state-of-the-art and baseline methods by a relatively large margin on this spontaneous database. According to the confusion matrix on SFEW database (Fig.~\ref{fig:confusion_total}),
disgust and fear are relatively difficult to recognize, whereas the recognition rate for happy is relatively high at 83.6\%. The primary reason is that capturing and labeling disgust, fear, and other less common expressions can be extremely difficult, particularly for the in-the-wild databases.

\subsubsection{The Real-world Affective Face Database (RAF-DB)}
Table~\ref{table:SFEW}, fourth column, displays the recognition rate for this relatively extensive ``in-the-wild'' database. Our approach achieves a recognition rate of 87.58\%, which is comparable to the state-of-the-art methods. Similar to the results from the SFEW database, the recognition rate for happy is extremely high at 94.3\%, whereas the recognition rates for disgust and fear are relatively low, as depicted in Fig.~\ref{fig:confusion_total}. Notably, recent methods such as SCN~\cite{Wang2020} or DMUE~\cite{She2021DiveIA} achieve high FER performance by addressing the issue of label ambiguity in the spontaneous databases. These methods relabel improperly labeled samples or adjust the learning focus based on the label distribution, in contrast to conventional perspectives that view manually annotated labels as 100\% accurate. We believe that by incorporating label ambiguity-resolving techniques into our methodology, FER performance can be further enhanced.

\subsection{Multitask-like FER Network}
\label{sec:multitask}
\begin{figure}
\centering
   \includegraphics[width=0.95\linewidth]{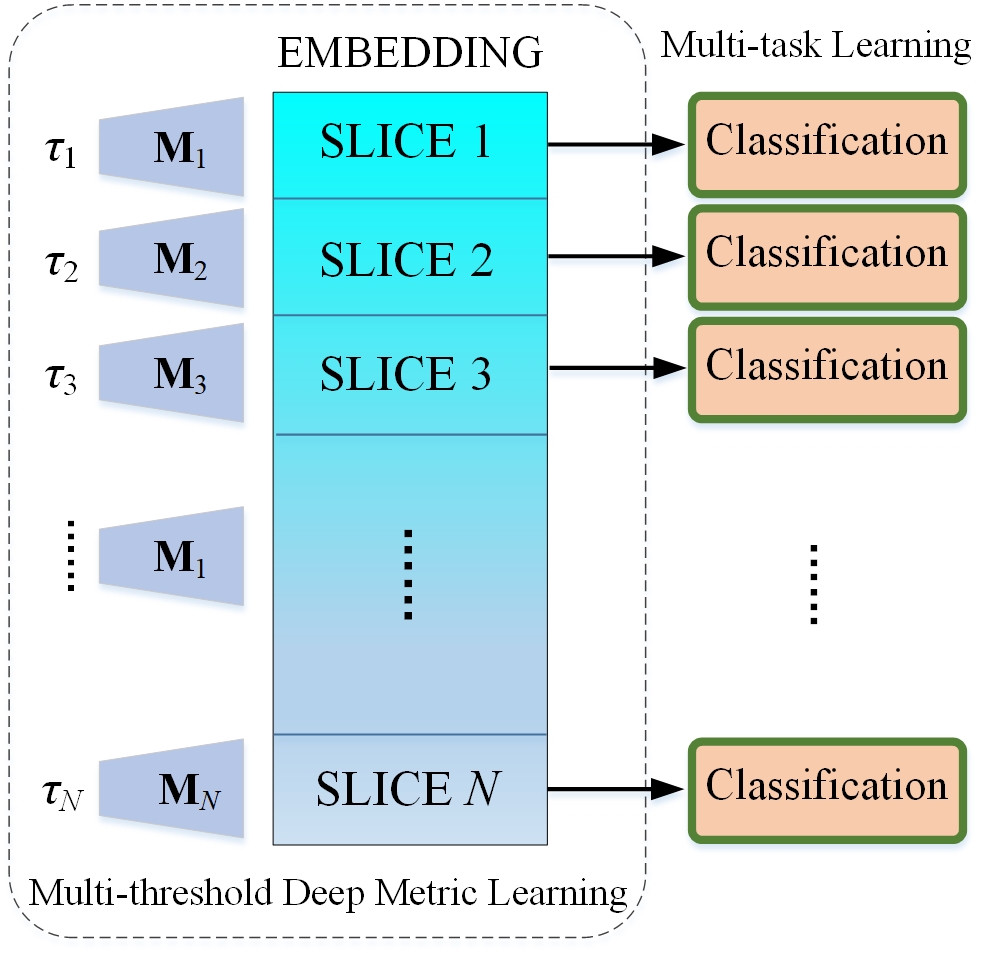}
   \caption{\label{fig:multi_task}
  A multitask-like FER network constructed with the multi-threshold deep metric learning technique, where each slice is treated as a distinct triplet loss learning network, followed by a classification layer.}
 \vspace{-2mm}
\end{figure}
Note that in the proposed multi-threshold deep metric learning technique, each embedding slice is formulated as a separate triplet loss learning network. As depicted in Fig.~\ref{fig:multi_task}, the multi-threshold deep metric learning technique can be used to construct a multitask-like FER network. In order to accomplish this, the last classification layer is removed from the proposed \mbox{Mul-DML model}, and each embedding slice is then followed by a classification layer, resulting in a multitask-like network in which each task performs a FER. A given input can be classified by an ensemble at inference time using majority voting. We refer to this type of network as \mbox{Mul-DML-Multask}. For comparison, we also build two baselines with the same structure as the \mbox{Mul-DML-Multask}: (1) \mbox{Mul-DML-Multask-Same} and (2) \mbox{CNN-Multask}. In the \mbox{Mul-DML-Multask-Same} model, all embedding slices are associated with the optimal threshold from Table~\ref{table:ablation_1Metric}, whereas the \mbox{CNN-Multask} model does not enforce triplet loss on embedding slices during training.

Clearly, the aforementioned baselines of the \mbox{S-Dual-Triplet} model and the \mbox{CNN} model represent the single-task versions of \mbox{Mul-DML-Multask-Same} and \mbox{CNN-Multask}, respectively. Both \mbox{Mul-DML-Multask-Same} and \mbox{CNN-Multask} improve the accuracy to some extent on each of the four databases relative to their respective single-task versions, as shown in Table~\ref{table:ablation_ensemble}. Intriguingly, the \mbox{Mul-DML-Multask} model, which employs different thresholds for each task, achieves accuracy comparable to that of the \mbox{Mul-DML-Multask-Same} model but is inferior to its \mbox{Mul-DML} counterpart. We believe it may be due to the majority-consistent voting-based ensemble strategy employed by \mbox{Mul-DML-Multask}, which requires consistency and seems to contradict the concept of diversity.
As a result, the various representation characteristics exhibited by distinct thresholds cannot be utilized to their full potential.

\begin{table}
\begin{center}
\caption{Performance comparison of multi-task models.}
 \label{table:ablation_ensemble}
\begin{tabular}{| L{85pt} | C{20pt} | C{20pt} | C{20pt} | C{30pt} |}
  \hline
   Models & CK+  &  MMI  &  SFEW  & RAF-DB \\
  \hline
  CNN & 93.99 & 71.25 & 54.58 & 86.63 \\
  S-Dual-Triplet & 96.51  & 74.33 &  57.34 & 85.63 \\
  Mul-DML & \textbf{98.49} &  \textbf{81.57} &  \textbf{60.09} & \textbf{87.58} \\
  \hline\hline
  CNN-Multask & 95.42 &  74.19 &  58.02 & 87.09 \\
  Mul-DML-Multask-Same & 96.76 &  75.91 & 58.48 & 87.14 \\
  Mul-DML-Multask & 96.59 &  76.16 &  58.49 & 87.22 \\
  \hline
\end{tabular}
\end{center}
\end{table}


\section{Conclusions}
In this paper, we present the end-to-end multi-threshold deep metric learning method for facial expression recognition.
We introduce a completely new viewpoint to triplet loss
learning by assuming that each threshold within a valid range
represents a unique representation characteristic that is distinct from
those of the other thresholds.
Rather than selecting a single optimal
threshold from a valid threshold range, we comprehensively sample
thresholds across the range, allowing the representation characteristics
manifested by thresholds within the range to be fully extracted
and leveraged for FER. Our method not only avoids
the difficult threshold validation but also vastly increases the
capacity of triplet loss learning to construct expression feature
representations.
Our method is straightforward to implement and adheres to the standard triplet loss learning paradigm, allowing it to be employed in the existing triplet-based deep metric learning frameworks in a plug-and-play fashion.
In addition, we demonstrate the problem of incomplete judgements inherent in the conventional triplet loss and then present a simple yet effective dual triplet loss that circumvents the problem and, as a result, achieves the rapid convergence of triplet loss learning.

The proposed method is extensively evaluated on both posed and spontaneous expression datasets, such as CK+, MMI, SFEW, and RAF-DB.
Compared to both state-of-the-art and baseline methods, our method achieves a consistent improvement in terms of FER accuracy, as demonstrated by the evaluation results.
Future efforts will concentrate on improving the model further and reducing its size.
In addition, we plan to apply the multi-threshold deep metric learning technique to other applications, such as image classification, image retrieval, and visual object recognition, where the triplet loss learning technique had been extensively used to learn a discriminative embedding.

\appendices
\section{}
\label{sec:validate_loss}

\begin{figure}[H]
\centering
   \includegraphics[width=0.9\linewidth]{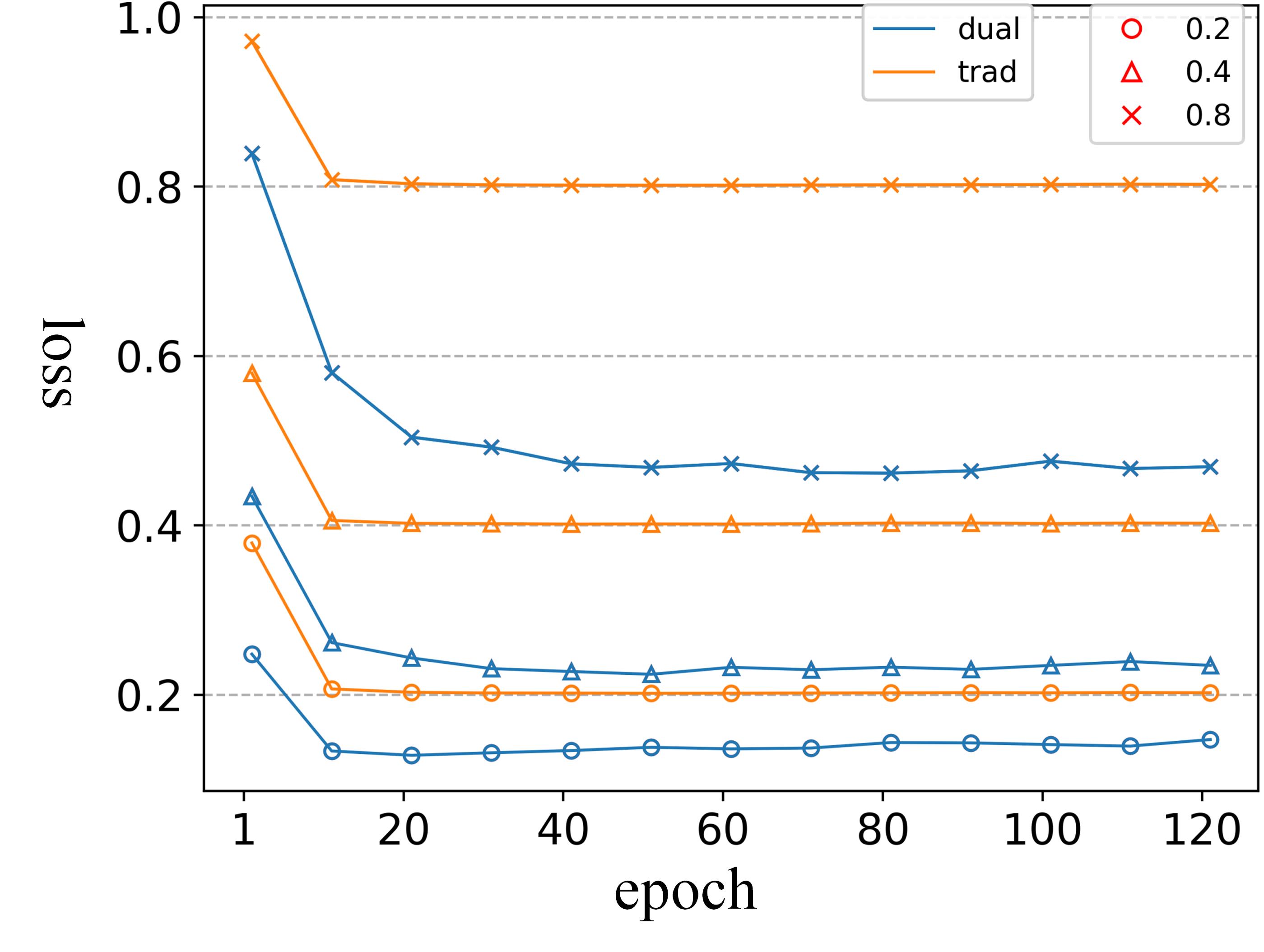}
   \caption{\label{fig:val_loss}
  The validation loss on the SFEW database, where three distinct thresholds ($0.2, 0.4, 0.8$) are employed for both the traditional triplet loss (trad) and the dual triplet loss (dual). }
\end{figure}

The validation loss of Fig.~\ref{fig:loss} in Section~\ref{sec:symmetric_triplet_loss} is depicted in Fig.~\ref{fig:val_loss}.
Similar results to the training loss in Fig.~\ref{fig:loss} can be observed in this figure, despite the existence of over-fitting. As shown in Table~\ref{table:SFEW}, the over-fitting is evidenced by the low recognition rate on the validation set of the SFEW database.


\section*{Acknowledgment}
The authors would like to thank Shuai Xing and Hao Chen for their help on the implementation and Yan Tian for his helpful discussion.
This research was partly funded by the ``Pioneer'' and ``Leading Goose'' R\&D Program of Zhejiang Province (2024C01167), the NSF of Zhejiang Province (LY21F020010), and NSF of China (61972353). Shen was in part supported by the FDCT grants 0102/2023/RIA2 and 0154/2022/A3. 

\ifCLASSOPTIONcaptionsoff
  \newpage
\fi



\bibliographystyle{IEEEtran}
\bibliography{main}


%





\end{document}